\documentclass[letterpaper, 10 pt, journal, twoside]{IEEEtran}

\IEEEoverridecommandlockouts                              

\title{Learning Solution Manifolds for Control Problems via Energy Minimization} 

\author{Miguel Zamora$^{1}$, Roi Poranne$^{{1}, {2}}$, and Stelian Coros$^{1}$

\thanks{This research was supported by the European Research Council (ERC) under the European Union’s Horizon 2020 research and innovation program (Grant No. 866480).}
\thanks{$^{1}$ The authors are with the Department of Computer Science, ETH, Zurich, Switzerland.
{\tt\footnotesize (miguel.zamora; roi.poranne; stelian.coros)@inf.ethz.ch}}%
\thanks{$^{2} $Roi Poranne is with the Department of Computer Science, University of Haifa, Haifa, Israel.
{\tt\footnotesize roiporanne@cs.haifa.ac.il}}%

}

\usepackage[dvipsnames]{xcolor}
\usepackage{colortbl}
\usepackage{comment}
\usepackage{amssymb}
\usepackage{graphicx}
\usepackage{physics}
\usepackage{amsmath}
\usepackage{mathtools}

\usepackage{algorithm,algorithmic}
\usepackage{booktabs}
\usepackage[hidelinks]{hyperref}

\usepackage[export]{adjustbox}

\ifdefined\showColumnMargins 
    \usepackage{showframe}
    \usepackage{etoolbox}
    
    \newlength\Fcolumnseprule
    \makeatletter
    \newcommand\ShowInterColumnFrame{
    \patchcmd{\@outputdblcol}
      {{\normalcolor\vrule \@width\columnseprule}}
      {\vrule \@width\Fcolumnseprule\hfil
        {\normalcolor\vrule \@width\columnseprule}
        \hfil\vrule \@width\Fcolumnseprule
      }
      {}
      {}
    }
    
    \makeatother
    \ShowInterColumnFrame
\fi

\usepackage{tikz}
\usetikzlibrary{automata,positioning}
\usetikzlibrary{backgrounds}
\usepackage{pgfplots}
\pgfplotsset{compat=newest}
\usetikzlibrary{plotmarks}
\usetikzlibrary{arrows.meta}
\usepgfplotslibrary{patchplots}
\usepackage{grffile}

\usepackage{wrapfig}
\usepackage{capt-of}
\usepackage{pgf}

\usepackage{subfig}
\usepackage{dblfloatfix}

 \usepackage{multirow}
\newcommand{\MZ}[1]{{\textcolor{black}{#1}}}  %

\newcommand{\E}{E}

\newcommand{\Hess}{\mathbf{H}}
\newcommand{\F}{\mathcal{F}}

\newcommand{\argmin}{\mathop{\mathrm{arg \hspace{0.1cm} min}}\limits}

\newcommand{\thetaVec}{\boldsymbol{\theta}}

\newcommand{\aTraj}{\Bar{ \boldsymbol{  a  } }}
\newcommand{\sTraj}{\Bar{ \boldsymbol{  s  } }}
\newcommand{\piTraj}{\Bar{\boldsymbol{\pi}}_{\thetaVec}}

\newcommand{\pVec}{\boldsymbol{p}}
\newcommand{\sVec}{\boldsymbol{s}}
\newcommand{\aVec}{\boldsymbol{a}}

\newcommand{\piPVec}[1][]{
\ifthenelse{\equal{#1}{}}{ \Bar{\boldsymbol{\pi}}_{\thetaVec}  }{ \Bar{\boldsymbol{\pi}}_{\thetaVec_{#1}} }
} 

\newcommand{\TVec}{\boldsymbol{\text{T}}} 

\def\borderColor{white}
\def\graphicsBorderColor{white}

\newcommand\convergenceWidth{0.125\textwidth} 

\newcommand\landscapeWidth{0.125\textwidth}
\newcommand{\plotLandscapes}[4]{
\begin{figure*}[#4]
\centering
\subfloat[]{
\resizebox{!}{\landscapeWidth}{%
      \includegraphics[width=\textwidth]{#1/SL}
    } 

}
\subfloat[]{
\resizebox{!}{\landscapeWidth}{%
      \includegraphics[width=\textwidth]{#1/Full_DAGGER}
    } 
}
\subfloat[]{
\resizebox{!}{\landscapeWidth}{%
      \includegraphics[width=\textwidth]{#1/Curry_Static}
    } 
}
\subfloat[]{
\resizebox{!}{\landscapeWidth}{%
      \includegraphics[width=\textwidth]{#1/Curry_MC}
    } 
}
\subfloat[]{
\resizebox{!}{\landscapeWidth}{%
      \includegraphics[width=\textwidth]{#1/Curry_DaggerDist}
    } 
}
\subfloat[]{
\resizebox{!}{\landscapeWidth}{%
      \includegraphics[width=\textwidth]{#1/Curry_DaggerDist_Reject}
    } 
}
\caption{#2. (a) SL. (b) Dagger. (c) Ours + static sampling. (d) Ours + Dynamic sampling. (e)  Ours + Incremental sampling. (g) Ours + Incremental + Rejection Sampling } \label{fig:plot-#3}
\vskip -0.2in
\end{figure*}

}
\newcommand\secondRowHeight{0.15\textwidth}

\newcommand{\plotConflicts}[4]{
\begin{figure*}[#4]
\centering
\subfloat{

\begin{tikzpicture}
    \node[rectangle,draw=\borderColor, minimum width = 0.08\textwidth, 
    minimum height = 0.15\textwidth] (r) at (0,0) {};
    
    \node[inner sep=0] at (r.center) { \includegraphics[width=0.15\textwidth, height=0.15\textwidth, keepaspectratio, cfbox=\graphicsBorderColor]{#1/bar} };
\end{tikzpicture}

}
\subfloat{
\begin{tikzpicture}
    \node[rectangle,draw=\borderColor, minimum width = 0.15\textwidth, 
    minimum height = 0.15\textwidth] (r) at (0,0) {};
    
    \node[inner sep=0] at (r.center) { \includegraphics[width=0.15\textwidth, height=0.15\textwidth, keepaspectratio, cfbox=\graphicsBorderColor]{#1/Energy_Curry_Static_512_Larger} };
\end{tikzpicture}
}
\subfloat{
\begin{tikzpicture}
    \node[rectangle,draw=\borderColor, minimum width = 0.15\textwidth, 
    minimum height = 0.15\textwidth] (r) at (0,0) {};
    
    \node[inner sep=0] at (r.center) { \includegraphics[width=0.15\textwidth, height=0.15\textwidth, keepaspectratio, cfbox=\graphicsBorderColor]{#1/Energy_Curry_Static_256} };
\end{tikzpicture}
}
\subfloat{
\begin{tikzpicture}
    \node[rectangle,draw=\borderColor, minimum width = 0.15\textwidth, 
    minimum height = 0.15\textwidth] (r) at (0,0) {};
    
    \node[inner sep=0] at (r.center) { \includegraphics[width=0.15\textwidth, height=0.15\textwidth, keepaspectratio, cfbox=\graphicsBorderColor]{#1/Energy_Curry_Static_128} };
\end{tikzpicture}
}
\subfloat{
\begin{tikzpicture}
    \node[rectangle,draw=\borderColor, minimum width = 0.15\textwidth, 
    minimum height = 0.15\textwidth] (r) at (0,0) {};
    
    \node[inner sep=0] at (r.center) { \includegraphics[width=0.15\textwidth, height=0.15\textwidth, keepaspectratio, cfbox=\graphicsBorderColor]{#1/Energy_Curry_Static_64} };
\end{tikzpicture}
}
\subfloat{
\begin{tikzpicture}
    \node[rectangle,draw=\borderColor, minimum width = 0.15\textwidth, 
    minimum height = 0.15\textwidth] (r) at (0,0) {};
    
    \node[inner sep=0] at (r.center) { \includegraphics[width=0.15\textwidth, height=0.15\textwidth, keepaspectratio, cfbox=\graphicsBorderColor]{#1/Energy_Curry_Static_32} };
\end{tikzpicture}
}
\quad
\subfloat{
\begin{tikzpicture}
    \node[rectangle,draw=\borderColor, minimum width = 0.08\textwidth, 
    minimum height = \secondRowHeight] (r) at (0,0) {};
    
    \node[inner sep=0] at (r.center) { \includegraphics[width=0.15\textwidth, height=\secondRowHeight, keepaspectratio, cfbox=\graphicsBorderColor]{#1/bar_knn} };
\end{tikzpicture}
}
\subfloat{
\begin{tikzpicture}
    \node[rectangle,draw=\borderColor, minimum width = 0.15\textwidth, 
    minimum height = \secondRowHeight] (r) at (0,0) {};
    
    \node[inner sep=0] at (r.center) { \includegraphics[width=0.15\textwidth, height=0.15\textwidth, keepaspectratio, cfbox=\graphicsBorderColor]{#1/Knn_Curry_Static_512_Larger} };
\end{tikzpicture}
}
\subfloat{
\begin{tikzpicture}
    \node[rectangle,draw=\borderColor, minimum width = 0.15\textwidth, 
    minimum height = \secondRowHeight] (r) at (0,0) {};
    
    \node[inner sep=0] at (r.center) { \includegraphics[width=0.15\textwidth, height=0.15\textwidth, keepaspectratio, cfbox=\graphicsBorderColor]{#1/Knn_Curry_Static_256} };
\end{tikzpicture}
}
\subfloat{
\begin{tikzpicture}
    \node[rectangle,draw=\borderColor, minimum width = 0.15\textwidth, 
    minimum height = \secondRowHeight] (r) at (0,0) {};
    
    \node[inner sep=0] at (r.center) { \includegraphics[width=0.15\textwidth, height=0.15\textwidth, keepaspectratio, cfbox=\graphicsBorderColor]{#1/Knn_Curry_Static_128} };
\end{tikzpicture}
}
\subfloat{
\begin{tikzpicture}
    \node[rectangle,draw=\borderColor, minimum width = 0.15\textwidth, 
    minimum height = \secondRowHeight] (r) at (0,0) {};
    
    \node[inner sep=0] at (r.center) { \includegraphics[width=0.15\textwidth, height=0.15\textwidth, keepaspectratio, cfbox=\graphicsBorderColor]{#1/Knn_Curry_Static_64} };
\end{tikzpicture}
}
\subfloat{
\begin{tikzpicture}
    \node[rectangle,draw=\borderColor, minimum width = 0.15\textwidth, 
    minimum height = \secondRowHeight] (r) at (0,0) {};
    
    \node[inner sep=0] at (r.center) { \includegraphics[width=0.15\textwidth, height=0.15\textwidth, keepaspectratio, cfbox=\graphicsBorderColor]{#1/Knn_Curry_Static_32} };
\end{tikzpicture}
}
\caption{#2} \label{#3}
\end{figure*}

}

\begin{document}

\maketitle

\begin{abstract}

A variety of control tasks such as inverse kinematics (IK), trajectory optimization (TO), and model predictive control (MPC) are commonly formulated as energy minimization problems. Numerical solutions to such problems are well-established. However, these are often too slow to be used directly in real-time applications. The alternative is to learn solution manifolds for control problems in an offline stage. Although this distillation process can be trivially formulated as a behavioral cloning (BC) problem, our experiments highlight a number of significant shortcomings arising due to incompatible local minima, interpolation artifacts, and insufficient coverage of the state space. In this paper, we propose an alternative to BC that is efficient and numerically robust. We formulate the learning of solution manifolds as a minimization of the energy terms of a control objective integrated over the space of problems of interest. We minimize this energy integral with a novel method that combines Monte Carlo-inspired adaptive sampling strategies with the derivatives used to solve individual instances of the control task. We evaluate the performance of our formulation on a series of robotic control problems of increasing complexity, and we highlight its benefits through comparisons against traditional methods such as behavioral cloning and Dataset aggregation (Dagger).

\end{abstract}



\section{Introduction}
\IEEEPARstart{O}{ptimization-based} control is one of the key components in the motion pipeline of some of the most advanced robots of today \cite{TO_anymal, goFetch, TO_atlas}. 

It can treat many highly nonlinear problems ranging from simple inverse kinematics to full-blown trajectory optimization.
However, optimization algorithms are still generally too demanding for real-time applications.
These applications require fast reaction times to input changes, and even simply changing an IK target for example, would require a slow re-optimization.

\begin{figure*}[t]
\vskip 0.2in	
	\centering
	\includegraphics[width=0.725\linewidth]{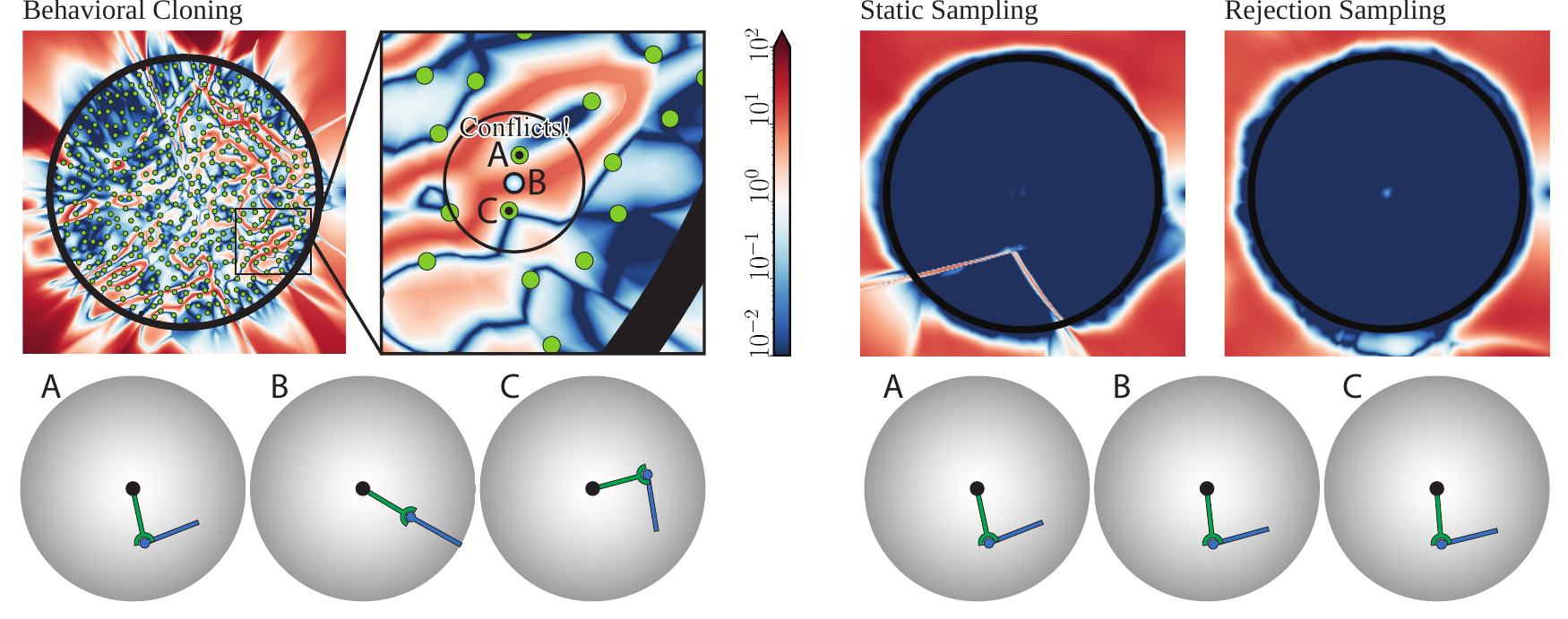}
	\caption{Comparison of objective landscapes (Eq. \ref{eq:ik-obj}) of a policy trained using Behavioral Cloning (BC) and our approaches. The dataset for BC was obtained by numerically solving 500 individual instances of the IK problem for a 2-Link mechanism with a maximum reach of 0.3m. The landscape was generated by evaluating the trained policy on 60k samples in a squared region of $0.3m \times 0.3m$. The black circle denotes the sampling region for training with a radius of 0.25m. The blow-up highlights regions with high energy values that are caused by conflicting samples, such as those denoted by A and C, where B is their interpolated mid-point. The bottom row shows the respective state. As can be seen, the state B indeed does not reach the target. In contrast, our sampling scheme exhibits no conflicts and generates low objectives. See the accompanying video for a visualization of the evolution of the energy landscapes for different methods as training progresses. \MZ{Note that the elbow-shaped conflicting regions depend on the random initialization of the policy.}}
	\label{fig:BC_energy_2Link}
\end{figure*}

Our goal is to make optimization-based control instantaneous using machine learning.
One possible approach is to first generate a dataset of solutions for different inputs using any optimization algorithm, and then train a neural network via supervised learning.
In other words, optimal solutions can be \emph{distilled} into high capacity functions approximators that can be queried in real time.
This is in fact a common approach known as \emph{behavioral cloning} (BC).
However, BC is known to produce policies that generalize poorly in the presence of conflicting samples in the dataset, which manifests as an erratic approximation landscape.
This is shown for a simple 2-link IK problem in Fig.~\ref{fig:BC_energy_2Link}.
Conflicting samples in turn are prone to happen when there are multiple solutions for the same task, i.e. \emph{multimodality}, and indeed, almost every IK target admits two solutions (Fig \ref{fig:BC_energy_2Link}).
As an alternative, we propose to formulate the distillation process as the minimization of the problem objective over the \emph{set of all inputs}, simultaneously.
Put differently, we minimize the \emph{integral} of the objective function over the entire input domain.
This approach does not rely on sampling solutions, and thus is more robust and results in more consistent solutions manifolds, as shown in Fig. \ref{fig:BC_energy_2Link}.

In this paper, we focus on \emph{trajectory-based policies} \cite{ILTaxonomy}, that is, policies that produce the entire sequence of actions at once.
This is in contrast to one-step policies that are evaluated iteratively, and generate only one action at a time.
We begin by recalling first-order and second-order update rules (Newton's method) necessary to minimize the objective for a \emph{single} problem instance.
We then show how a \emph{sample} of the inputs can be optimized in a reformulation of the problem as a BC problem, but one that can leverage the same first and second order information. 
We examine the performance of our approach compared to standard BC, and investigate different sampling and dataset aggregation strategies.
We then tackle multimodal objectives that might generate conflicting data, by introducing detect-and-reject mechanism, which exhibits smoother solution manifolds and a better performance overall.
Throughout the paper we use a 2-Link planar robot as a guiding example, and demonstrate the generality of our method in the results section.



\section{Related work}

Leveraging optimization-based control to train neural network policies has been of great interest for both the robotics and the learning communities \cite{Mordatch, montgomery2016guided}. 
Behavioral cloning, an instance of imitation learning \cite{ILTaxonomy}, is one of the simplest ways to train policies, using expert demonstrations generated by well-established optimization algorithms \cite{BC_with_TO, kahn2017plato}.
However, its performance can be hindered by data conflicts that often occur due to multimodality (see e.g. \cite{overcoming_conflicts, pinneri2021extracting} and Fig. \ref{fig:BC_energy_2Link} ).

Our main focus is training trajectory policies for problems that can be formulated as one-shot decision making problems, such as simple IK or more involved kinematic trajectory optimization.
We ground our formulation on gradient-based optimization of an energy function using second order information. 
Other gradient-based formulations to train policies are presented in \cite{supertrack, mora2021pods} in a differentiable simulation framework, and in \cite{2014-cgps, montgomery2016guided} under the umbrella of guided policy search methods.
However, their focus is on training reactive one-step policies.
Other approaches that are concerned with trajectory policies include \cite{TS_PI_LearningMovementPrimitives, TS_PI_DynamicMovementPrimitives, TS_PI_ProbabilisticMovementPrimitives2} in the context of Motion Primitives, 
\cite{TS_PI_PolicyGradient} where the policy  defines directly a trajectory distribution, \cite{iscen2019policies} learns policies that modulate trajectory generators for robot locomotion, and
\cite{learn_solution_manifold} trains a neural network to output weighting matrices to mix different basis functions that represent full trajectories.

As mentioned, one of the challenges arise due to multimodality.
Generative Adversarial imitation learning (GAIL) \cite{gail} enables learning multimodal solutions, but \cite{imitation_divergence_min} argues that GAIL and similar approaches still struggle with the problem of interpolating modes.
The paper also suggests that learning one type of solutions should suffice for robotics applications, and we adopt this viewpoint in this paper as well.
Other approaches include \cite{Keep_Doing_What_Worked} which deals with conflicts by introducing an advantage weighted behavior model to regularize the policy, \cite{pinneri2021extracting} by using an adaptive auxiliary cost weighting, and \cite{imitation_divergence_min} by minimizing an f-divergence metric.
Our approach is to detect and remove conflicting samples from the training set, as discussed in the following sections.

\MZ{
Further recent approaches that display good performance in the presence of multimodal data are Soft-Q Imitation Learning (SQIL) \cite{sqil}, which reformulates BC as an reinforcement learning problem by assigning rewards when the policy matches a demonstrated action in a demonstrated state,  and Implicit Behavioral Cloning (IBC) \cite{ibc},  which we discuss in the results section.
Our approach can also be interpreted as an algorithm for imitation learning using a converging supervisor \cite{converging_supervisor}.
However, our focus is on trajectory-polices in contrast to the reactive policies used in \cite{converging_supervisor}.
Finally, we share the goal of accelerating optimization-based control with \cite{accelerate_grasp}. 
But, their focus is on kinematic planning, while we apply our method to both kinematic and dynamic problems.
}
\section{Method}
In the following section we first recall the standard second-order gradient-based approach, e.g. Newton's method, used to solve individual problem instances. 
We then formulate the problem of learning the space of optimal solutions as the minimization of an integrated objective over an input domain, which can be approximated by a finite sum.
While this problem can be optimized directly, we show that it can be solved via BC steps that use targets derived from first and second-order information (Converging supervisor). 
We show how these targets can enjoy standard optimization techniques such as line-search or the Gauss-Newton approximation.
We conclude this section by presenting different sampling a strategies to discretize the integration domain, and introduce a method to detect and reject conflicting samples in a dataset.

\subsection{Solving individual problem instances}\label{sec:individual_problems}
Consider an optimal control problem of the form
\begin{alignat}{4}
&  \aTraj^* & =  & \,\,  \argmin_{\aTraj}  \E( \sTraj(\aTraj, \pVec), \aTraj, \pVec),   \label{eq:singleProblemInstance}
\end{alignat}  
where $\E$ is an energy function, $\sTraj := [\sVec_1, \dots, \sVec_t]$ and $\aTraj := [\aVec_0, \dots, \aVec_{t-1}]$ are the state and control trajectories, and $\pVec$ is a set of input parameters.
The notation $\sTraj(\aTraj, \pVec)$ indicates that the state trajectory depends on the actions and the inputs.
For the simple IK problem we discuss in this section, $\E$ is defined by
\begin{equation}\label{eq:ik-obj}
\E = \|\F(\aTraj) - \pVec\|^2 + w_0\|\aTraj - \aTraj_{ref}\|^2
\end{equation}
where $\aTraj$ represents the joint angles, $\pVec$ is the desired end-effector position, $\F$ is the forward-kinematics map, and $\aTraj_{ref}$ serves as a regularization term with a small weight $w_0$.
Note that $\pVec$ can include information about a particular initial state, a target position, or a physical property of the system.
The action trajectory $\aTraj$ can be optimized with update rules, based on gradient descent and Newton's method:
\begin{itemize}
    \item First order update rule
    \begin{equation} \label{eq:TS_1st_update}
      \aTraj  \,\,  \xleftarrow{}   \,\, \aTraj - \alpha_{\aTraj} \biggl[ \dv{ \E( \sTraj(\aTraj), \aTraj ) }{\aTraj} \biggr]
    \end{equation}
    \item Second order update rule
\begin{equation} \label{eq:TS_2nd_update}
      \aTraj  \,\,  \xleftarrow{}   \,\, \aTraj - \alpha_{\aTraj} {\biggl[ \dv[2]{\E( \sTraj(\aTraj), \aTraj )}{\aTraj}  \biggr]}^{-1} \biggl[ \dv{ \E( \sTraj(\aTraj), \aTraj ) }{\aTraj}  \biggr]
\end{equation}
\end{itemize}
where we neglect $\pVec$ for conciseness. The terms $\dv{ E( \sTraj(\aTraj), \aTraj ) }{\aTraj}$ and $\dv[2]{E( \sTraj(\aTraj), \aTraj )}{\aTraj}$ are the gradient and Hessian, and $\alpha_{\aTraj}$ is a learning rate, which can be determined using a line search procedure to ensure monotonic improvements.
We note that in practice, it is preferable to use the Gauss-Newton approximation of the Hessian since it is positive definite, which guarantees a descent direction \cite{puppetMaster,num_opt}.

\subsection{Learning the optimal solution space}
We formulate the learning process as a minimization of the integral of an energy function over a space of input parameters $P$ as follows:
\begin{alignat}{4}
&  \thetaVec^* &  = & \,\,  \argmin_{\thetaVec}  \int _{ P }^{  }{ E( \sTraj(\piTraj, \pVec), \piTraj, \pVec )  \,\, d\pVec}, \label{eq:mighty_Energy_Integral}
\end{alignat} 
where $\piTraj:= [\aVec_0, \dots, \aVec_{t-1}]$ is a neural network policy that produces a one-shot sequence of actions given a set of input parameters $\pVec$ and a set of weights $\thetaVec$.
The integral can be approximated via \emph{Monte-Carlo integration}, using a finite sum of random samples:
\begin{alignat}{4}
&  \thetaVec^* &  \approx & \,\,  \argmin_{\thetaVec}  \frac{1}{M} \sum _{m=1}^{M}{ E( \sTraj(\piTraj, \pVec^{m}), \piTraj, \pVec^{m} ) }.  \label{eq:discreteEnergy}
\end{alignat} 
We immediately obtain a first order update rule for the sum,
\begin{equation}\label{eq:PS_1st_update}
      \thetaVec  \,\,  \xleftarrow{}   \,\, \thetaVec - \alpha_{\thetaVec} \frac{1}{M} \sum _{m=1}^{M}{   \dv{ E( \sTraj(\piTraj, \pVec^{m}), \piTraj, \pVec^{m} ) }{\thetaVec}}.
\end{equation}
The summands can be evaluated using the chain-rule:
\begin{equation}
     \dv{ E( \sTraj(\piTraj), \piTraj, \pVec^{m}) }{\thetaVec}  \,\, =  \,\, \dv{ E( \sTraj(\piTraj), \piTraj, \pVec^{m} ) }{\piTraj} \dv{ \piTraj }{\thetaVec}.
\end{equation}

We can interpret this optimization in the parameter space $\thetaVec$ as BC by considering the supervised learning loss
%
\begin{equation}
     L   \,\,  = \,\, \frac{1}{M} \sum _{m=1}^{M}{\frac{1}{2} || \piTraj^{m} - \TVec^m||}^2,  \nonumber
\end{equation}    
where $\piTraj^{m} \coloneqq \piTraj(\pVec^{m})$ and $\TVec^m$ is a target policy.
An update rule for $\thetaVec$ that minimizes $L$ is
\begin{equation}
     \thetaVec  \,\, \xleftarrow{}  \,\, \thetaVec - \alpha_{L} \sum _{m=1}^{M} (\piTraj^{m} - \TVec^m)^{T} \dv{ \piTraj^{m} }{\thetaVec}  . \label{eq:BC_1st_order}
\end{equation}
If we replace $\TVec^m$ by 
\begin{equation} \label{eq:1st_order_target}
     \TVec^m  \,\, = \,\, \piTraj^{m} - \alpha_{\aTraj} \dv{ E( \sTraj(\piTraj^{m}), \piTraj^{m} ) }{\piTraj^{m}}
\end{equation}
and substitute into the supervised update rule \eqref{eq:BC_1st_order} we obtain
\begin{equation*}
      \thetaVec  \,\, \xleftarrow{}  \,\, \thetaVec - \alpha_{L} \alpha_{\aVec} \frac{1}{M} \sum _{m=1}^{M}  \dv{ E( \sVec(\piTraj^{m}), \piTraj^{m} ) }{\piTraj^{m}} \dv{ \piTraj^{m} }{\thetaVec}  ,
\end{equation*}
which matches the first-order update rule for energy minimization \eqref{eq:PS_1st_update} when setting $\alpha_{L}$ appropriately.
Note that while it is counterproductive to apply a line-search procedure to \eqref{eq:PS_1st_update}, we can use line-search on $\alpha_{\aTraj}$ to guarantee $E( \sTraj(\TVec^M), \TVec^{m} ) < E( \sTraj(\piTraj^{m}), \piTraj^{m} )$, which stabilizes the training process.

Taking this idea further, we define a second-order target
\begin{equation}
     \TVec^m  \,\,  = \,\, \piTraj^{m} - \alpha_{\aVec} { \Hess }^{-1} \dv{ E( \sTraj(\piTraj^{m}), \piTraj^{m} ) }{\piTraj^{m}} \label{eq:2nd_order_target}
\end{equation}
where $\Hess =\dv[2]{ E( \sTraj(\piTraj^{m}), \piTraj^{m} ) }{ (\piTraj^{m}) } $ is the Hessian.
Substituting the second-order targets in \eqref{eq:BC_1st_order} yields the update rule
\begin{equation} \label{eq:Curry_update_rule}
     \thetaVec  \,\, \xleftarrow{}  \,\, \thetaVec - \alpha_{L} \alpha_{\aVec} \sum _{m=1}^{M} \dv{ E( \sVec(\piTraj^{m}), \piTraj^{m} ) }{\piTraj^{m}}  { \Hess }^{-1} \dv{ \piTraj^{m} }{\thetaVec}.
\end{equation}
This can also be viewed as an update rule that minimizes the weighted squared error loss.
\begin{equation}\label{eq:weighted_loss}
      L  \,\,   =  \,\, \frac{1}{M} \sum _{m=1}^{M}{\frac{1}{2}(\piTraj^{m} - \TVec^m)^{T} W^m (\piTraj^{m} - \TVec^m) },
\end{equation}    
with $W^m = { \Hess }^{-1}$.
In practice, it is better to use the Gauss-Newton approximation.
Algorithm \ref{alg:EM_Curry} presents our suggested approach together with a sampling strategy described below.

\vspace{3mm}
\begin{algorithm}
   \caption{Energy Minimization}
   \label{alg:EM_Curry}
\begin{algorithmic}
   \FOR{$k=1$ {\bfseries to} $K$}
   \STATE Sample $M_k$ input parameters $\pVec^{m}$
   
   \STATE // Define converging targets
   \FOR{$m=1$ {\bfseries to} $M_k$}
   \STATE $\TVec^m   =  \piTraj^{m} - \alpha_{\aVec} { \Hess }^{-1} \dv{ E( \sTraj(\piTraj^{m}), \piTraj^{m} ) }{\piTraj^{m}}^{T}$
   \ENDFOR
   
   \STATE // Find non-conflicting samples 
   \STATE $ Dataset =  ( (\pVec^0, \TVec^0), \dots,  (\pVec^{NC}, \TVec^{NC}) )$
   
   \STATE // Improvement in PS
   \FOR{$n=1$ {\bfseries to} $N$}
   \STATE $\thetaVec  \xleftarrow{}  \thetaVec - \alpha_{L} \sum _{d=1}^{NC}\biggl[ (\piTraj^{d} - \TVec^d)^{T} \dv{ \piTraj^{d} }{\thetaVec}  \biggr]^{T} $
   \ENDFOR
   \ENDFOR
\end{algorithmic}
\end{algorithm}

\subsection{Strategic Sampling for policy learning}
Using \eqref{eq:Curry_update_rule} with a static set of samples from $P$, already exhibits better behaved energy landscapes compared to BC (Fig. \ref{fig:BC_energy_2Link}).
We attribute this to the targets being dynamic and monotonically improving.
Additionally, BC cannot resolve conflicting data, which might be present during the entire training process. 
While conflicts may still exist in dynamic targets, they tend to vanish during optimization.
Nevertheless, a fixed set of \emph{samples} is still limited, and some conflicts might remain due to multimodality of the solution space.
Indeed, the choice of samples in \eqref{eq:discreteEnergy} has a significant impact in both performance and generalization.
In the following we propose strategies to mitigate that.

\begin{figure}[t]
\centering
\begin{tikzpicture}
\node[inner sep=0pt] (pend) at (0,0)    {\includegraphics[width = 0.7\columnwidth]{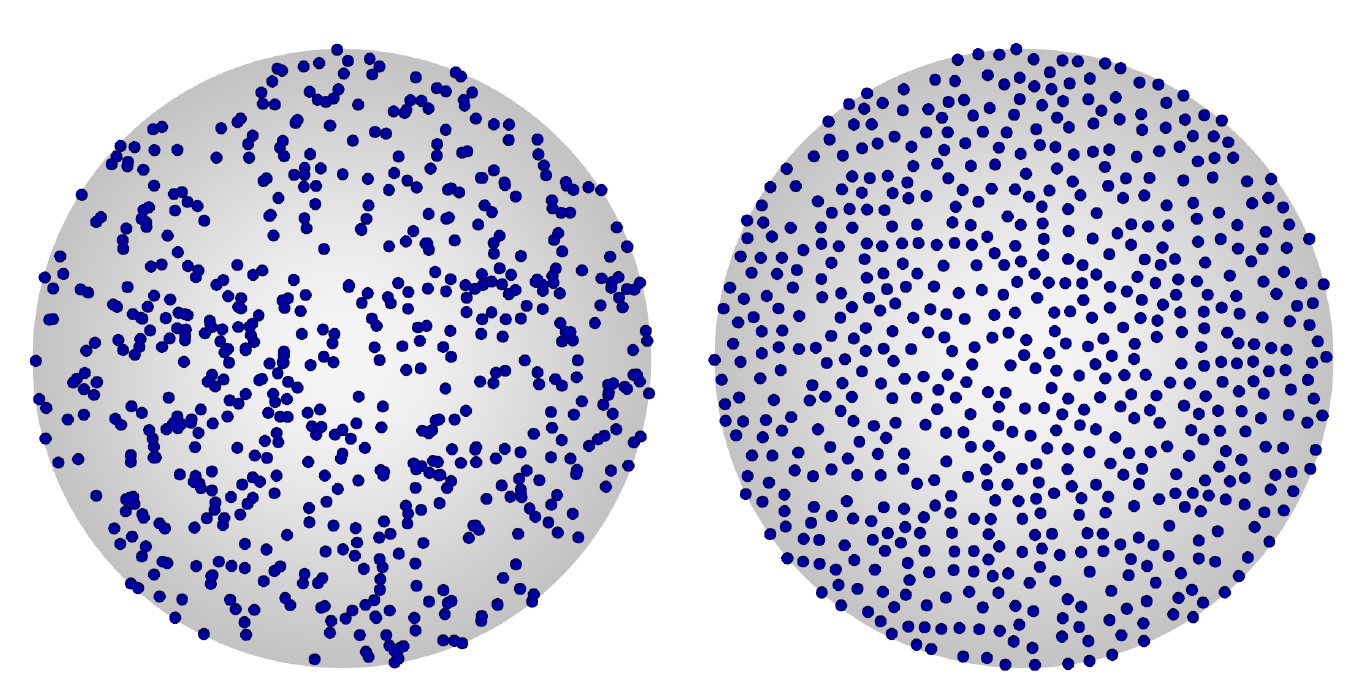} }; 

\end{tikzpicture} 
    \caption{\textbf{Left to right;} Uniform, and Poisson-Disk sampling} 
    \label{fig:unif_vs_pds}
\end{figure}

\noindent\textbf{Static Sampling.} The samples $\pVec^{m}$ are selected once and are kept fixed.
Instead of the common uniform or Gaussian sampling, we propose to employ Poisson-disk sampling (PDS) \cite{poisson_disk_sampling}. 
PDS spreads samples more evenly (Fig. \ref{fig:unif_vs_pds}), which results in a better coverage and easy detection of conflicting data as explained below.
    
\noindent\textbf{Dynamic Sampling.} There is in fact no reason to keep the parameter sample $\pVec^{m}$ fixed at every iteration.
To the contrary, resampling allows for better coverage of the input domain, for the same computational budget.
This can be viewed as SGP with batches taken from a continuous distribution.

\noindent\textbf{Incremental sampling.} 
We can also gradually increase the size of the input sample set.
Specifically, we begin with a seed sample that is chosen by sampling a batch of inputs and selecting the sample that produces the lowest energy with the default weights of the policy.
We then sample again increasing the number of samples, but keep only the ones that are close to the seed sample. Once a maximal number of samples is reached, we stop expending the domain.

We found that this sampling method can induce uniform solution manifolds.
We attribute such behavior to a bootstraping effect that the overall training process presents.
To illustrate this, consider the case of a single sample.
After one iteration, the output of the policy will match the first target that leads to a lower energy region.
Using a second sample that is close to the seed sample to query the policy is likely to result in a target that is non-conflicting with the target of the seed sample.
In contrast to Dagger methods, we are not aggregating samples labeled with fully optimal solutions, but we grow a dataset of samples labeled with converging 2nd order targets that change dynamically at every iteration. 
    
\noindent\textbf{Rejection sampling}
As mentioned, interpolating two conflicting data points and their corresponding labels leads to high energy regions.
We can encourage non-conflicting datasets by searching and removing samples that when averaged with their nearest neighbours result in high energy (Fig \ref{fig:BC_energy_2Link}).
Algorithm \ref{alg:MonitorAndReject} describes the process, which relies on a \emph{rejection} measure $D$.
The measure can be the energy itself,
but for interpretability, it can also be any other function that describes discrepancy between two averaged samples. 
For the IK examples we used the position error as a metric, which allowed us to set the threshold $\epsilon$ in an intuitive way.
Furthermore, using PDS we can compute the Poisson disk radius, which denotes half the distance between the closest pair of samples, to set the search radius $r$ in Algorithm \ref{alg:MonitorAndReject} equals to twice the Poisson disk radius.

\plotConflicts{plots/Landscapes/XY_2L_Conflicts}{ \textbf{Top row}: Detecting conflicts using different testing set sizes (512, 256, 128, 64, 32) for the energy landscapes of policies trained using our method with static samples. Purple and cyan markers denote conflicting and non-conflicting samples respectively. \textbf{Bottom row}: Position error of the samples averaged with the neighbours within twice the PSD radius. Horizontal axis enumerates samples in the testing set. The red line is the average test error of the interpolated samples. }{fig:knn_sizes}{ht}

\begin{figure*}[ht]
\centering
\begin{tikzpicture}
\node[inner sep=0pt] (pend) at (0,0)    {\includegraphics[width = 0.90\textwidth]{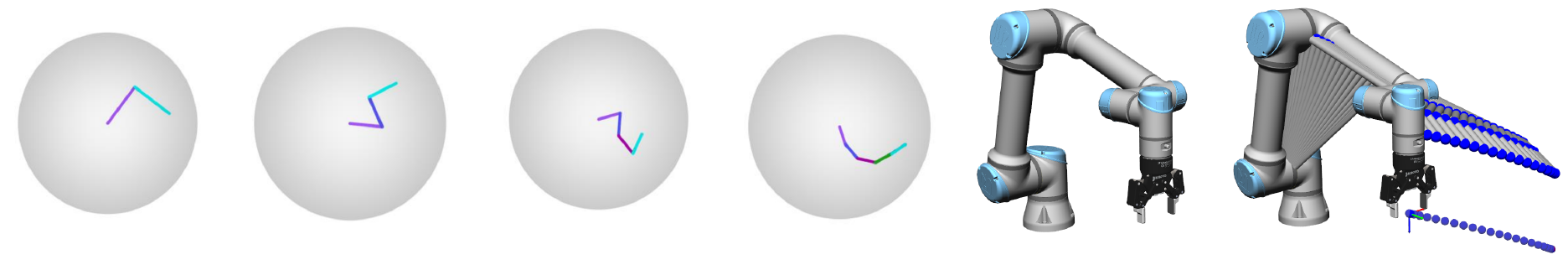} }; 
\end{tikzpicture} 
    \caption{\textbf{Left to right:} Planar IK for n-link serial robots $n=2 \text{ to } 5$. IK for UR5, Kinematic TO for UR5} 
    \label{fig:RoboticProblems}
\vskip -0.2in
\end{figure*}

\vspace{3.0mm}
\begin{algorithm}
   \caption{Detect and reject conflicts}
   \label{alg:MonitorAndReject}
\begin{algorithmic}
   \STATE {\bfseries Input:} $ Dataset =  ( (\pVec^0, \TVec^0), \dots,  (\pVec^{M}, \TVec^{M}) )$
   \FOR{$m=0$ {\bfseries to} $M$}
   \STATE Find neighbours of $\pVec^{m}$ within a radius of $r$
   \STATE Interpolate all neighbours to get $\pVec^{m_{avg}}$
   \STATE Interpolate associated targets $\TVec^{m_{avg}}$
   \STATE Evaluate and Store metric $D^m(\pVec^{m_{avg}}, \TVec^{m_{avg}})$
   \ENDFOR
   \STATE Compute average $D^{m_{avg}}$
   \FOR{$m=0$ {\bfseries to} $M$}
   \IF{ $D^{m} >  D^{m_{avg}} + \epsilon$ }
   \STATE Reject sample $m$ and all its neighbouring samples.
   \ENDIF
   \ENDFOR
\end{algorithmic}
\end{algorithm}

Note that for the $Dataset$ in Algorithm \ref{alg:MonitorAndReject}, $\TVec^m$ is a converging target and not a final solution obtained by performing $K$ steps of energy minimization.
In practice, we apply this rejection mechanism at every iteration once the targets $\TVec^m$ have been updated, as noted in Algorithm \ref{alg:EM_Curry}, which result in a uniform energy landscape (Fig. \ref{fig:BC_energy_2Link}). 

To illustrate the effectiveness of the conflict detection mechanism, we applied Algorithm \ref{alg:MonitorAndReject} on testing sets with different sizes, for a policy trained using our method with static sampling for the 2-Link IK problem.
Figure \ref{fig:knn_sizes} shows how even for narrow regions in the space of input parameters, we can detect conflicting data using a relatively sparse number of samples, which we consider an important feature as sampling a higher dimensional space has a sparser nature.
Furthermore, notice that samples around the origin are also highlighted as conflicting samples, which is reasonable as there exist infinite solutions to the IK problem for the origin. 

\section{Experimental setups}\label{subsec:robotic-problems}
In this section we describe in more detail the series of kinematic problems that we used to evaluate our approach. Details concerning the network architectures that we used are presented in Appendix \ref{appdx:net_arch}.

\noindent\textbf{Planar IK for serial n-link mechanisms:} \\
The goal for this family of problems is to learn the IK solutions of planar serial mechanisms with different number of links (Fig. \ref{fig:RoboticProblems}, top row).
The inputs $\pVec$ to the network are 2D end effector targets, and the output is a set of $n$ joint angles $\piPVec(\pVec)$.
This experiment shows the performance on increasingly complex systems.
Note that as the number of link $n$ increases, so does the dimensionality of possible IK solutions, which in turn increases the chances for conflicts.

Considering the iterative nature of the optimization approach, we denote the current output of the policy as $\piPVec[i](\pVec)$. The energy function is then:
\begin{multline*}
 E( \sVec, \piPVec[i](\pVec), \pVec )  \,\,  =   \,\, w_0 ||\sVec_{0} - \sVec_{Pos}( \piPVec[i](\pVec)  ) ||^2 \,\, + \hfill \\[3pt]
  \hfill w_1 || \piPVec[i](\pVec) - \aVec_{ref} ||^2 \,\, + \,\, w_2 || \piPVec[i](\pVec) - \piPVec[i-1](\pVec) ||^2 
\end{multline*}
%
where $\sVec_{Pos}( \piPVec[i](\pVec) )$ represents the Forward Kinematics mapping from joint angles $\piPVec[i](\pVec) $ to Cartesian coordinates $\sVec_{Pos}$, and $\aVec_{ref}$ is a reference set of joint angles. 
    
\noindent\textbf{IK for UR5 serial robot with self collision:} \\
Here we consider a UR5 robot, with a fixed gripper orientation pointing downwards, and a collision avoidance term following  \cite{SimonD_elastic} (see Appendix \ref{appdx:other_objs}).
The inputs are 3D Cartesian target and the outputs are sets of 6 joint angles. The energy function is then:
\begin{multline*}
E( \sVec, \piPVec[i](\pVec), \pVec )  \,\,  =  \,\, E_{joint \,\, limits}(\aVec_i) \,\,  + \,\,  E_{collision}(\aVec_i) \,\, + \hfill \\[3pt]
   \hfill  w_0 || \piPVec[i](\pVec) - \aVec_{ref} ||^2 \,\, +  \,\, w_1 || \piPVec[i](\pVec) - \piPVec[i-1](\pVec) ||^2 \,\,\, + \\[3pt]
   \hspace{6mm} w_2 ||\sVec_{0} - \sVec_{Pos}( \piPVec[i](\pVec)  ) ||^2 \,\, + \,\, w_3 || \sVec_{{Rot}_{0}}  -  \sVec_{Rot}( \piPVec[i](\pVec) ) ||^2 
\end{multline*}

\begin{figure*}[h]
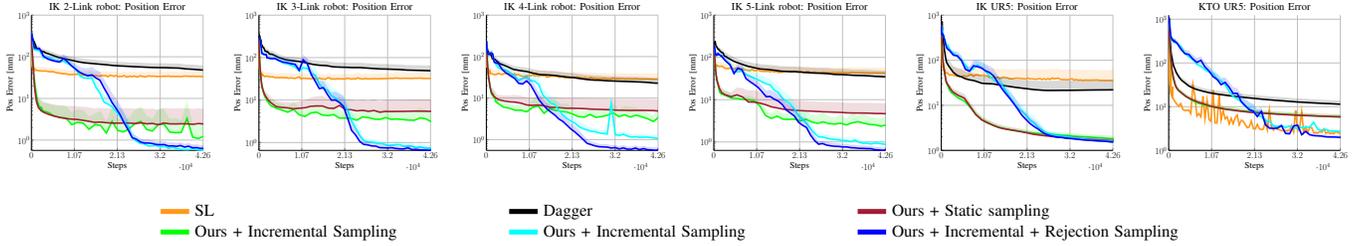

\centering
\subfloat{
\resizebox{!}{\convergenceWidth}{%
      \input{plots/PosError/Energy_IK_XY_2L_PosError}
    } 

}
\subfloat{
\resizebox{!}{\convergenceWidth}{%
      \input{plots/PosError/Energy_IK_XY_3L_PosError}
    } 
}
\subfloat{
\resizebox{!}{\convergenceWidth}{%
      \input{plots/PosError/Energy_IK_XY_4L_PosError}
    } 
}
\subfloat{
\resizebox{!}{\convergenceWidth}{%
      \input{plots/PosError/Energy_IK_XY_5L_PosError}
    } 
}
\subfloat{
\resizebox{!}{\convergenceWidth}{%
      \input{plots/PosError/Energy_IK_UR5_PosError}
    } 
}
\subfloat{
\resizebox{!}{\convergenceWidth}{%
      \input{plots/PosError/Energy_KTO_UR5_PosError}
    } 
}
\quad
\subfloat{
\resizebox{!}{0.035\textwidth}{%
      \definecolor{color_SL}{rgb}{1.0000,    0.5961,    0.0902 }
\definecolor{color_Full_Dagger}{rgb}{0     ,         0,         0 }%
\definecolor{color_MC}{rgb}{0,   1,   0 }%

\definecolor{color_Static}{rgb}{ 0.6314, 0.1098, 0.1961 }

\definecolor{color_IncrementalMC_Metric}{rgb}{0     ,         1,         1 }%
\definecolor{color_IncrementalMC_Metric_Reject}{rgb}{0     ,         0,         1 }

\begin{tikzpicture} 
    \begin{axis}[%
    hide axis,
    xmin=0,
    xmax=1,
    ymin=0,
    ymax=0.4,
    legend style={legend columns=3, legend cell align=center, draw=none, text width=6.5cm}
    ]
    \addlegendimage{color_SL,mark options={solid},line width=3.0pt}
    \addlegendentry{SL}; 
    
    \addlegendimage{color_Full_Dagger,mark options={solid}, line width=3.0pt}
    \addlegendentry{Dagger}; 

    \addlegendimage{color_Static,mark options={solid}, line width=3.0pt}
    \addlegendentry{Ours + Static sampling}; 
    
    \addlegendimage{color_MC,mark options={solid}, line width=3.0pt}
    \addlegendentry{Ours + Incremental Sampling};
    
    
    \addlegendimage{color_IncrementalMC_Metric,mark options={solid}, line width=3.0pt}
    \addlegendentry{Ours + Incremental Sampling };
    
    \addlegendimage{color_IncrementalMC_Metric_Reject,mark options={solid}, line width=3.0pt}
    \addlegendentry{Ours + Incremental + Rejection Sampling};
    \end{axis}
\end{tikzpicture}
    } 
}
\caption{Comparison of position error on a testing set for different training methods. Testing sets for the planar n-Link problems and the UR5 problems contained 500 and 2000 samples respectively. (For clarity we only plot the upper side of the shaded region that represents the standard deviation). } \label{fig:plot-PosError}
\vskip -0.2in
\end{figure*}

\plotLandscapes{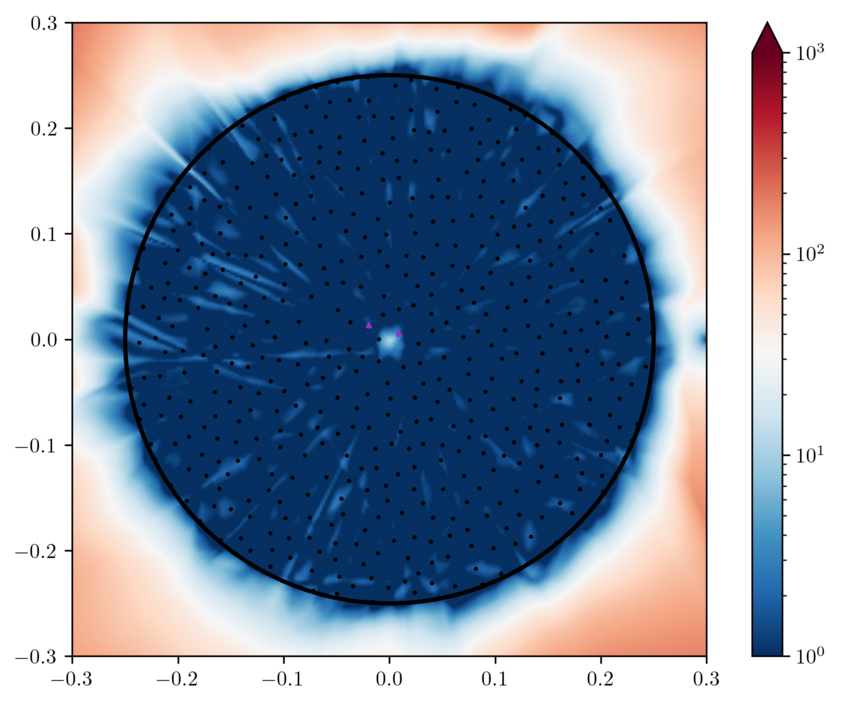}{Position Error Landscape for a 5-Link planar robot}{XY_5L_PosErrorLandscape}{h}
\noindent\textbf{Kinematic trajectory optimization:} \\
Here we learn feasible trajectories that travel from one point to another.
As above, the gripper orientation remains fixed, and we include additional objectives to ensure the smoothness and feasibility of the trajectory.
In this case, the one-shot policy $\piPVec[i](\pVec)$ is defined as a sequence of joint angles $\aVec_j$, that is $\piPVec[i](\pVec) = [{\aVec}^{T}_{0} \dots  {\aVec}^{T}_{t-1} ]^{T} $. Moreover, the input parameters $\pVec = [{\aVec}^{T}_{ref} \,\, {\sVec}^{T}_{ {Pos}_{t}} ]^T $ contain a set of initial joint angles ${\aVec}^{T}_{ref}$ that describe the initial robot configuration and serve also as a regularizer state, and $\sVec_{{Pos}_{t}}$ a target Cartesian position. The energy function is then:
\vspace{1.5mm}
\begin{multline*}
E( \sVec, \piPVec[i](\pVec), \pVec )  \,\, =  \,\, E_{feasibility}( \sVec, \piPVec[i](\pVec), \pVec ) \,\, + \hspace{45mm} \, \\[3pt]
\hfill \,\,  E_{smoothness}( \sVec, \piPVec[i](\pVec), \pVec ) \,\, + \,\,  E_{collision}( \sVec, \piPVec[i](\pVec), \pVec ) \,\, + \\[3pt]
\hfill \,\, w_0 ||\sVec_{{Pos}_{t}} - \sVec_{Pos}( \aVec_0 ) ||^2  + w_1 || \sVec_{{Rot}_{t}}  -  \sVec_{Rot}( \aVec_0 ) ||^2 \,\, + \\[3pt]
\hfill \,\, w_2 \sum _{j=0}^{t-1} || \aVec_j - \aVec_{ref} ||^2 \,\, + \,\, w_3 || \piPVec[i](\pVec) - \piPVec[i-1](\pVec) ||^2 
\end{multline*}
\vspace{1.5mm}

The objectives $E_{feasibility}$, $E_{smoothness}$, $E_{collision} $, presented in more detail in Appendix \ref{appdx:other_objs}, encourage the feasibility in terms of joint limits at the position, velocity, acceleration, and jerk levels, the smoothness of the control trajectory in terms of small velocities, accelerations and jerks, and the feasibility in terms of collisions.

\section{Results}
We evaluate the performance of our method with different sampling strategies against BC and Dagger. For the sake of interpretability and despite the fact that we minimize energy functions that weight different objectives, we report the average position error for the different training methods that we compare (Figure  \ref{fig:plot-PosError}). We used training and testing sets of 500 and 2000 samples for the planar n-Link problems,  and the problems involving the UR5, respectively.

\MZ{To illustrate the fairness of our comparisons, consider that in the planar IK problems, we used 500 input samples for BC (the computational cost of input samples is almost negligible), and then we applied the Gauss-Newton update rule (Eq. \ref{eq:TS_2nd_update}) $\mathbf{K}$ times, for each input sample, to generate a fully labeled dataset, with an associated computational budget of $\mathbf{500K}$. For our method with static or dynamic sampling, it's straightforward to see that doing $\mathbf{K}$  iterations of Algorithm \ref{alg:EM_Curry} and given the same number of input samples, the computation budget is roughly the same as the one associated to BC.
For the case of incremental sampling or incremental + rejection sampling, we keep track of the number of times that Eq. \ref{eq:2nd_order_target} is called to make sure it does not exceed the computational budget of $\mathbf{500K}$.
Furthermore, while Dagger is traditionally used in the context of reactive policies, we applied it to our trajectory-based policy setting. At every Dagger iteration we get 1 new input sample using PDS, we obtain a trajectory from $\piTraj$ and we relabel such trajectory by applying $\mathbf{K}$ times the Gauss-Newton update rule (Eq. \ref{eq:TS_2nd_update}).
We aggregate the relabeled trajectories and we train  $\piTraj$ on the the aggregated dataset. We iterate until the computational budget that was allocated for BC is reached.
}

The axis of Figure  \ref{fig:plot-PosError} represents only the gradient steps effectively performed on $\thetaVec$, as we consider it is most standard.  
Our method with rejection sampling shows a slower initial convergence than BC, which can be explained by the fact that BC has already access from the beginning to both a set of samples that discretize the entire domain of interest and full labels obtained via optimization based control.
In contrast, the rejection sampling method includes only a small region of the space of input parameters during the early stages, while the test set spans the entire domain of integration.
Furthermore, we show the corresponding position error landscape of the learned policies for the 5-Link example (Figure \ref{fig:plot-XY_5L_PosErrorLandscape}) to highlight how our method results in lower and more uniform landscapes. 
Note that, while for some tasks, our method with incremental MC sampling achieves uniform landscapes, using rejection sampling achieves even lower error with lower variance, in general.


\subsection{Warm-starting optimization}
Using the learned policies in a similar fashion to \cite{accelerate_grasp},  and warm-starting the KTO problem for the UR5 robot (Section \ref{sec:individual_problems}), on a test set of 1000 input samples, yields a reduction of optimization time from  47ms to 3ms, on average, with a position and orientation error of 0.05 mm and $0.05^{\circ}$, while the trained policy alone evaluated on the same test set, achieves an average error of 2.04 mm and $0.06^{\circ}$.

\subsection{Broader class of systems}
\begin{figure}[h]
\vskip 0.2in
\centering
\includegraphics[width=0.925\columnwidth]{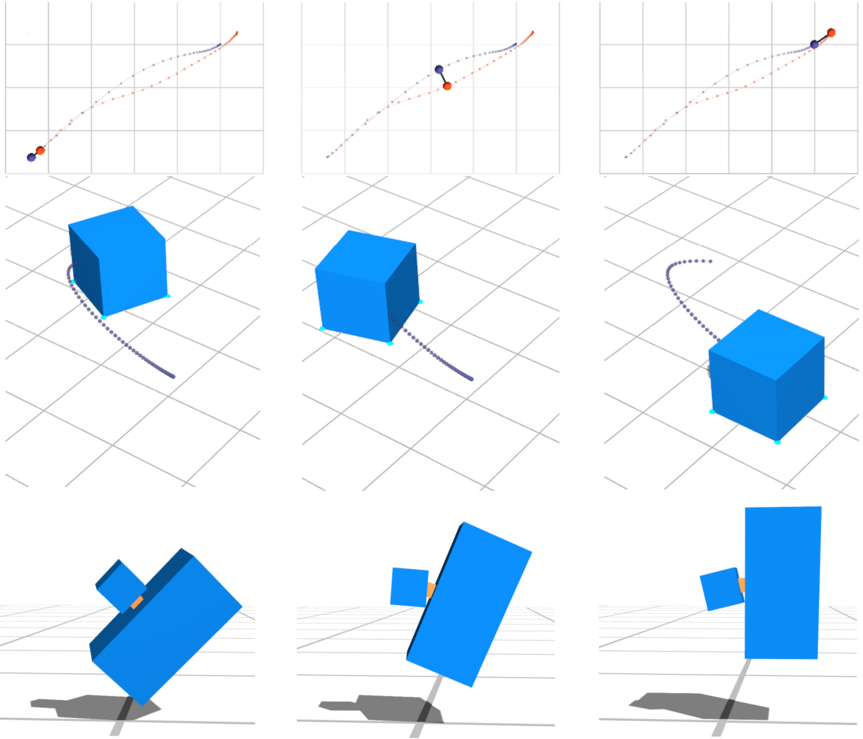}
\caption{\textbf{Top Row:} Dragging a point mass (purple) towards the origin. \textbf{Middle Row:} Repositioning and reorienting a rigid body box sliding over a frictional surface.
\textbf{Bottom Row:} Reorienting a rigid body box with a flexible attachment.}
\label{fig:dynamic-experiments}

\vskip -0.2in
\end{figure}

The optimization scheme presented in the Method section is very general. When applied to kinematic trajectory optimization problems, it can be seen as direct transcription method. Furthermore, when the map $\sTraj(\aTraj, \pVec)$ represents a dynamical system, such formulation can be seen as a direct shooting method. To exemplify such generality we applied our method to the problems shown in Fig \ref{fig:dynamic-experiments}. The goal of the first two examples is to drag a point mass or rigid body towards the origin over a surface with a differentiable friction model as implemented by \cite{add_moritz}. For the point mass example the controls represent the position of the handle that is attached to the point mass via a spring. For the rigid body box, in the second example, the controls are the forces applied to the center of mass and a torque applied along the vertical axis to achieve a target orientation. The goal of the third experiment is to reorient a box that is connected to a rigid body via a flexible attachment, and the controls are the position and orientation of the smaller rigid body.

\subsection{Multimodal behaviour}
We trained Implicit Behavioral Cloning (IBC) \cite{ibc} to solve the planar n-Link IK problems. IBC used a dataset of 500 samples (500 input samples each fully labeled with a possible  IK solution). We trained IBC using an InfoNCE loss with batches of 256 positive samples and 256 random uniform negative samples per positive sample. At inference time we used a derivative free optimization method with 16384 samples to optimize the energy landscape the IBC model represents.
While IBC preserves multimodality, our method performs better in terms of position error and compute time as shown in  table \ref{table:ibc_comparisson}. Additional experiments show that the performance of IBC improves with access to larger datasets. 
    
\begin{table}[h]
\caption{Performance on a test set of 512 samples }
\label{table:ibc_comparisson}
\vskip 0.15in
\begin{center}
\begin{small}
\begin{sc}
\begin{tabular}{llcccc}
\toprule
                      & \multicolumn{2}{c}{Ours} & \multicolumn{2}{c}{IBC}     \\
\midrule
\multicolumn{1}{c}{IK} & \multicolumn{1}{c}{\begin{tabular}[c]{@{}c@{}}Error \\ {[}mm{]}\end{tabular}} & \multicolumn{1}{c}{\begin{tabular}[c]{@{}c@{}}Time \\ {[}min:sec{]}\end{tabular}} & \multicolumn{1}{c}{\begin{tabular}[c]{@{}c@{}}Error \\ {[}mm{]}\end{tabular}} & \multicolumn{1}{c}{\begin{tabular}[c]{@{}c@{}}Time\\ {[}min:sec{]}\end{tabular}} \\
\midrule
2-Link & 0.63           &       3:09           & 2.2            & 82  \\
3-Link & 0.65           &       3:03           & 7.1            & 104 \\
4-Link & 0.56           &       2:33           & 26.5           & 108 \\
5-Link & 0.64           &       3:01           & 76.6           & 103 \\
\bottomrule
\end{tabular}
\end{sc}
\end{small}
\end{center}
\vskip -0.1in
\end{table}
\section{Conclusion and future work}

We presented a gradient-based framework to learn trajectory-policies via the minimization of an energy integral over a domain of interest. 
We showed how to reformulate such minimization into a sequence of small BC problems by using first and second order targets. 
Furthermore, we investigated different sampling strategies to discretize the domain of integration and we introduced a simple mechanism to detect and reject conflicting samples. 
Such strategies allowed us to learn consistent solution manifolds despite the multimodality of the solution space. 
However, the sample strategies that worked best rely on KNN which might prevent the scalability of the approach to larger datasets. 
Extending our method to enable learning multiple solutions manifolds is also another exciting research direction.
Further ways to leverage our mechanism to detect conflicts should be explored. 
An interesting direction would be to use the detection mechanism to cluster different solutions. 
This can enable extending our work into a multimodal setting. 
Furthermore, detecting conflicting areas in the domain can also be used to indicate regions that required sampling more densely.

Further sampling strategies should be explored. 
Extending our work to use weighted Poisson disk sampling will enable coarse and fine sampling, in different regions of the domain, depending on the nature of problem.
We believe better rejections mechanisms should be explored. 
In this work, we reject all the samples associated to a conflict. 
Such rejection rule, however, might be too conservative. 
Finally, we believe this work, with all its exciting future venues of research, sets a good starting point to enable robust and instantaneous queries from well-stablished optimization-based control algorithms.






\section*{APPENDIX}

\subsection{Additional objectives} \label{appdx:other_objs}
\begin{wrapfigure}{r}{0.375\columnwidth}
\vskip -0.3in
\centering
\includegraphics[width=0.375\columnwidth]{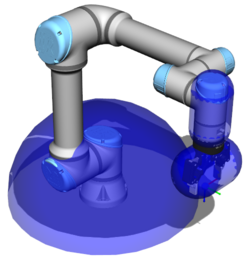}
\caption{Collision \\ primitives}
\label{fig:collision-primitives}
\end{wrapfigure} 
\noindent\textbf{Collision} \\       
Following the approach presented in \cite{SimonD_elastic}, we define a set of collision primitives using spheres and capsules, as shown in figure \ref{fig:collision-primitives}. The energy is then defined using unilateral barrier function \cite{jim_barrier}  to  ensure that the minimum distance between collision primitives is larger than a safety threshold. \\
\noindent\textbf{Feasibility}
\begin{multline*}
    E_{feasibility}( \piPVec[i](\pVec), \pVec ) \,\, = \,\,  E_{joint \,\, pos \,\, limits}(\piPVec[i](\pVec)) \,\, +  \\[3pt]
     \hspace{5mm} \,\,  E_{joint \,\, vel \,\, limits} (\piPVec[i](\pVec))  +   E_{joint \,\, acc \,\, limits} (\piPVec[i](\pVec)) \,\, +  \\[3pt]
     \hspace{5mm} \,\,  E_{joint \,\, jerk \,\, limits}(\piPVec[i](\pVec))
\end{multline*}
The energy associated to joint limits is implemented using bilateral barrier functions as presented in \cite{jim_barrier}. \\
\noindent\textbf{Smoothness} 
    \begin{multline*}
    E_{smoothness}( \piPVec[i](\pVec), \pVec ) \,\, =  \,\,   E_{small \,\, vel}(\piPVec[i](\pVec), \pVec) \,\, + \\[3pt]
    E_{small \,\, acc }(\piPVec[i](\pVec), \pVec) \,\, + E_{small \,\, jerk }(\piPVec[i](\pVec), \pVec)
    \end{multline*}
    where each individual term is simply the squared norm of the corresponding quantity computed using a second order backward finite difference \cite{backwardFD} and assuming the robot stands still in its initial configuration $\aVec_{ref}$ e.g.
    \begin{multline*}
    E_{small \,\,  vel}( \piPVec[i](\pVec), \pVec) \,\, =   \,\, || \frac{3}{2}\aVec_{0} - 2\aVec_{ref} + \frac{1}{2}\aVec_{ref}||^2 \, + \hfill  \\[3pt]
    \hspace{8mm} || \frac{3}{2}\aVec_{1} - 2\aVec_{0} \,\, + \frac{1}{2}\aVec_{ref}||^2 + \sum _{j=2}^{t-1} || \frac{3}{2}\aVec_{j} - 2\aVec_{j-1} + \frac{1}{2}\aVec_{j-2}||^2
    \end{multline*}


\subsection{Network Architecture}\label{appdx:net_arch}
For the experiments described in this paper, we used two hidden layers with relu activation functions and 512 nodes, and an output layer with tanh activations. 

    



    

As we deal with periodic variables for the joint angles of a robot, we encoded the output of the network such that each each angle $\theta_i$ is represented by tuple $(\sin{\theta_i}, \cos{\theta_i})$. As done by \cite{ik_distal_learning}. A theoretical justification can be found in \cite{Bishop_Pattern_Recognition} (Pages 105-110). We found such encoding to alleviate some of the problems related to conflicting data, and to improve performance as well. Table \ref{table:problem-dims} summarizes the dimensionality of the problems described in section \ref{subsec:robotic-problems}. Note that because of the encoding we used, the output dimension doubles.
    
\begin{table}[h]
\caption{Dimensionality of problems.}
\label{table:problem-dims}
\vskip 0.15in
\begin{center}
\begin{small}
\begin{sc}
\begin{tabular}{lccr}
\toprule
Problem & Input Dim & Output Dim \\
\midrule
IK n-Link mechanism    & 2 & 2*n  \\
IK UR5                 & 3 & 2*6  \\
KTO UR5                & 9 & 2*6*30 = 180 \\
\bottomrule
\end{tabular}
\end{sc}
\end{small}
\end{center}
\vskip -0.1in
\end{table}

\vspace{-3.0mm}
\section*{ACKNOWLEDGMENT}
We thank Vittorio Megaro for the initial explorations of energy minimization, Ramon Witschi for the plotting tools, and Kiran Doshi for the help running comparisons.
\vspace{-0.5mm}

\bibliographystyle{IEEEtran}
\bibliography{IEEEabrv,root.bib}

\end{document}